\documentclass[journal,twoside,web]{ieeecolor}
\let\labelindent\relax
\usepackage{enumitem}
\usepackage{generic}
\usepackage{cite}
\usepackage{amsmath,amssymb,amsfonts}
\usepackage{booktabs}
\providecommand{\refname}{References}

\newcommand{\mystat}[2]{\ensuremath{#1 \,\pm\, {\scriptstyle #2}}}

\usepackage{tabularx}
\usepackage[table]{xcolor}
\usepackage{graphicx}
\usepackage{algorithm,algorithmic}
\usepackage{multirow}
\usepackage{hyperref}
\hypersetup{hidelinks=true}
\usepackage{textcomp}
\def\BibTeX{{\rm B\kern-.05em{\sc i\kern-.025em b}\kern-.08em
    T\kern-.1667em\lower.7ex\hbox{E}\kern-.125emX}}
\begin{document}

\title{Bridging Modalities on the Cortex: Surface-based MRI to PET Translation \\ with a Diffusion Bridge}

\author{Yitong Li, Alexandra Samoylova, Fabian Bongratz, Timo Grimmer, \\Dennis M. Hedderich, Igor Yakushev, and Christian Wachinger
\thanks{Y. Li, A. Samoylova, F. Bongratz, and C. Wachinger are with the Technical University of Munich (TUM), TUM University Hospital, and Munich Center for Machine Learning (MCML), Germany}
\thanks{T. Grimmer is with the Department of Neurology, TUM University Hospital, Germany.}
\thanks{D. M. Hedderich is with the Department of Neuroradiology, TUM University Hospital, Germany.}
\thanks{I. Yakushev is with the Department of Nuclear Medicine, TUM University Hospital, Germany.}
\thanks{This work was supported by the German Research Foundation (DFG), MCML, and the DAAD programme Konrad Zuse Schools of Excellence in Artificial Intelligence, sponsored by the Federal Ministry of Research, Technology, and Space.}
\thanks{Yitong Li and Alexandra Samoylova contributed equally to this work.} 
\thanks{Email: \{yi\_tong.li, alexandra.samoylova, fabi.bongratz, t.grimmer, dennis.hedderich, igor.yakushev, christian.wachinger\}@tum.de}}

\maketitle

\begin{abstract}
Cortical hypometabolism measured by Fluorodeoxyglucose Positron Emission Tomography (FDG-PET) is a highly sensitive biomarker for dementia diagnosis. However, high costs, radiation exposure, and limited accessibility constrain its clinical utility. 
While cross-modal synthesis from Magnetic Resonance Imaging (MRI) offers a promising alternative, existing volumetric generation methods do not explicitly account for the highly folded cortical geometry, where disease-related patterns predominantly reside.
To address this, we introduce a novel surface-based diffusion bridge framework \emph{DB-SUiT} for MRI-to-PET translation that operates natively on the cortical manifold. A conditional Spherical U-shaped vision Transformer (SUiT) is specifically designed to model the intricate cross-modal relationships while preserving surface topology. It combines spherical convolutional encoders for multi-scale surface feature extraction with bottleneck Transformers to capture long-range spatial dependencies, while incorporating demographic and subcortical conditions to refine the synthesis. 
Evaluated on two datasets, including subjects with different dementia types, DB-SUiT demonstrates high-fidelity synthesis that substantially outperforms other baselines. 
In automated dementia classification, synthesized PET surfaces improve performance over MRI by 14.2\% and PET volumes by 11.3\%, approaching the performance of real PET surfaces. 
In a blinded reader study, synthetic PET achieved 85.5\% diagnostic accuracy, compared with 75.8\% for MRI and 95.2\% for real PET. This further demonstrates cross-cohort and cross-pathology generalization, as the model was evaluated without retraining on an external cohort that included a dementia subtype not represented during training.
Our code is available at \url{https://github.com/ai-med/DB-SUiT}.
\end{abstract}

\begin{IEEEkeywords}
MRI, PET, Surface, Translation, Diffusion models, dementia diagnosis.
\end{IEEEkeywords}

\section{Introduction}
\label{sec:introduction}

\begin{figure*}[t]
    \centering
    \includegraphics[width=0.9\linewidth]{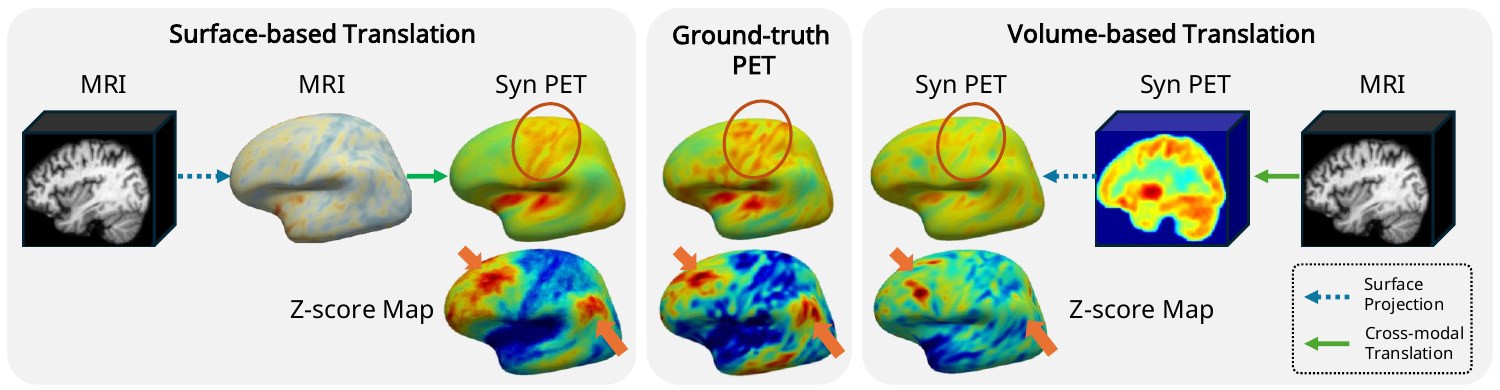}
    \caption{
Comparison of surface- and volume-based MRI-to-PET translation. 
Left: Surface-based translation maps MRI-derived cortical features directly to synthetic (Syn) PET on the cortical manifold, followed by normative z-score calculation. Center: Ground-truth PET and its z-score map serve as reference. Right: Volume-based translation generates synthetic PET in voxel space before surface projection. The highlighted regions show that surface-based translation more closely reproduces the localized metabolic abnormalities in the ground truth, particularly in the temporoparietal and frontal lobes. In contrast, volume-based translation produces smoother and less pronounced hypometabolic patterns.
    }
    \label{fig:teaser}
\end{figure*}
Characteristic patterns of cortical atrophy and hypometabolism are key biomarkers for dementia diagnosis and are largely expressed across the cerebral cortex~\cite{strom2022cortical,minoshima1995diagnostic}. Their analysis in volumetric images is limited by the highly folded cortical geometry and by partial-volume effects at tissue boundaries~\cite{dale1999cortical,thomas2011importance}. 
Surface-based representations address these limitations by explicitly modeling the cortical manifold and enabling anatomically aligned analysis of spatially distributed disease patterns. 
They are therefore widely used to quantify structural MRI properties such as cortical thickness and curvature~\cite{fischl2000measuring,tustison2014large}.
Surface-based visualization also has a long tradition in FDG-PET. Since the introduction of three-dimensional stereotactic surface projection by Minoshima et al.~\cite{minoshima1995diagnostic}, cortical surface displays have become an established tool for visualizing characteristic patterns of hypometabolism. More recent methods extend this principle by projecting PET measurements onto detailed cortical meshes extracted from MRI ~\cite{marcoux2018automated,park2006cortical,tan2018quantitative}, supporting spatially precise and anatomically corresponding analyses across the cortical surface.



Although PET often provides more sensitive diagnostic signals than structural MRI in dementia~\cite{shivamurthy2015brain,minoshima202218f}, its clinical adoption is limited by high costs, ionizing radiation, and restricted availability~\cite{pet_cost,huang2009whole}. Structural MRI, in contrast, is noninvasive and routinely acquired in clinical workflows~\cite{pet_ov_mri}. This discrepancy has motivated growing interest in cross-modal translation from MRI to PET~\cite{gandalf,lin2021bidirectional,Li2024pasta,chen2025multi,li2025diffusion,li2026translating}. Existing approaches, however, perform this translation exclusively in the volumetric domain. They therefore model cortical signals within a Euclidean voxel grid that does not explicitly represent the topology and highly folded geometry of the cortical sheet.

By projecting PET measurements onto MRI-derived cortical meshes, these recent methods place structural and metabolic information in vertex-wise anatomical correspondence on the same subject-specific cortical surface. This creates the opportunity to formulate MRI-to-PET generation directly on the cortical manifold, rather than first synthesizing a volumetric PET image and projecting it onto the surface afterward. As illustrated in Fig.~\ref{fig:teaser}, conventional approaches perform the cross-modal mapping in voxel space before surface projection. A direct surface-based formulation could instead exploit the shared cortical representation throughout generation, potentially preserving the spatial organization and regional specificity of disease-related metabolic patterns more effectively.
Building on this motivation, we propose DB-SUiT, a surface-based framework that formulates MRI-to-PET translation as a diffusion bridge directly on the cortical manifold. The framework models the conditional transformation from structural MRI features to metabolic PET signals while preserving their vertex-wise anatomical correspondence. As its backbone, we introduce the Spherical U-shaped vision Transformer (SUiT), which combines spherical ResNets for multiscale surface feature extraction with bottleneck Transformers for capturing long-range dependencies across the cortical sheet. DB-SUiT further incorporates subject-specific demographic information and subcortical volumes as complementary conditioning variables. Our main contributions are:
\begin{enumerate}
\item We introduce the first MRI-to-PET translation framework that generates metabolic signals directly on the cortical manifold using anatomically corresponding MRI and PET surface representations.
\item We formulate surface-based MRI-to-PET translation as a diffusion bridge process and introduce SUiT, which combines spherical convolutions with Transformers to model local and long-range cross-modal relationships while supporting deterministic generation.
\item Extensive evaluations on two datasets demonstrate superior generation quality and significantly improved automated diagnostic performance. 
\item A blinded clinical reader study on an independent test dataset demonstrates higher diagnostic accuracy with synthetic PET than structural MRI.
\end{enumerate}


\section{Related Work}

\subsection{Surface-Based Brain Analysis}
Surface-based brain representations support geometrically informed analyses of the cerebral cortex and have been widely adopted for cortical reconstruction~\cite{dale1999cortical} and thickness estimation~\cite{fischl2000measuring,tustison2014large}.
For PET, surface projection methods such as three-dimensional stereotactic surface projection (3D-SSP) and cortical mesh-based PET quantification have been used to improve visualization, regional analysis, and clinical interpretation of cortical metabolism~\cite{minoshima1995diagnostic,park2006cortical,tan2018quantitative,marcoux2018automated}. 
In parallel, advances in geometric deep learning have enabled neural networks to operate directly on cortical surfaces by extending operations to non-Euclidean manifolds. Spherical U-Net adapts convolutional encoder-decoder operations to spherical cortical representations~\cite{zhao2019spherical}, while Surface Vision Transformers and their multiscale extensions model long-range dependencies across surface patches using self-attention~\cite{dahan2022surface,dahan2024multiscale}. 
These methods enable surface representations as a powerful domain for deep learning-based cortical analysis.

\subsection{MRI-to-PET Cross-Modal Translation}
Early volumetric MRI-to-PET translation methods relied on generative adversarial networks (GANs): GANDALF~\cite{gandalf} integrated diagnosis-driven discriminator-adaptive loss fine-tuning for end-to-end synthesis; Lin et al.~\cite{lin2021bidirectional} proposed a 3D reversible GAN to learn bidirectional MRI-PET mappings, and Hu et al.~\cite{hu2021bidirectional} introduced BMGAN with bidirectional image- and latent-space mapping for brain MR-to-PET synthesis. More recently, diffusion-based approaches have advanced this field: PASTA~\cite{Li2024pasta,li2026translating} introduced a pathology-aware dual-arm conditional diffusion framework with cycle exchange consistency; Xie et al.~\cite{xie2024synpet} used a Joint Diffusion Attention Model (JDAM) for PET synthesis from high-field and ultra-high-field MR images; Chen et al.~\cite{chen2025multi} combined multi-view diffusion synthesis with downstream classifiers; and SiM2P~\cite{li2025diffusion} demonstrated clinical-grade PET simulation via a 3D diffusion bridge validated through a blinded reader study, with local-adapt for site-specific translation.
Diffusion bridge models provide a principled framework for paired image translation by explicitly learning the stochastic trajectory between source and target domains~\cite{bbdm,peluchetti2023diffusion,zhou24ddbm}, making them well-suited for MRI-to-PET synthesis where MRI provides strong anatomical constraints, and PET provides complementary metabolic information. However, existing MRI-to-PET synthesis methods remain volumetric and therefore do not explicitly exploit the intrinsic geometry of cortical metabolic patterns.

\section{Proposed Method}

\subsection{Multimodal Cortical Representations}

We represent cortical measurements on a spherical icosahedral mesh $\mathcal{M} = (\mathcal{V}, \mathcal{E})$ at subdivision level~5 (ico5), yielding $N =10,242$ vertices $\mathcal{V}=\{{v_i}\}^N_{i=1}$ per hemisphere.
The signals from different modalities are represented as vertex-wise feature maps.
Structural MRI is encoded as $\mathbf{x}_m \in \mathbb{R}^{N \times 1}$, denoting the cortical thickness value at each vertex $v_i$ obtained from FreeSurfer~\cite{fischl2012freesurfer} reconstruction. 
Cortical thickness is used as the primary bridge source input as it is the most established and robust surface biomarker of neurodegeneration.
FDG-PET is mapped to the same manifold as $\mathbf{x}_p \in \mathbb{R}^{N \times 1}$, representing vertex-wise normalized cortical glucose metabolism.
We further extract a subject-level tabular feature $\mathbf{h} \in \mathbb{R}^{38}$, comprising age, sex, and 36 subcortical volumetric measures\footnote{Details are in \url{https://github.com/ai-med/DB-SUiT}.}, 
e.g., hippocampal and ventricle volumes. 
These features provide demographic and non-cortical structural context that complements the local cortical thickness signal.

\begin{figure*}[t]
    \centering
    \includegraphics[width=0.9\linewidth]{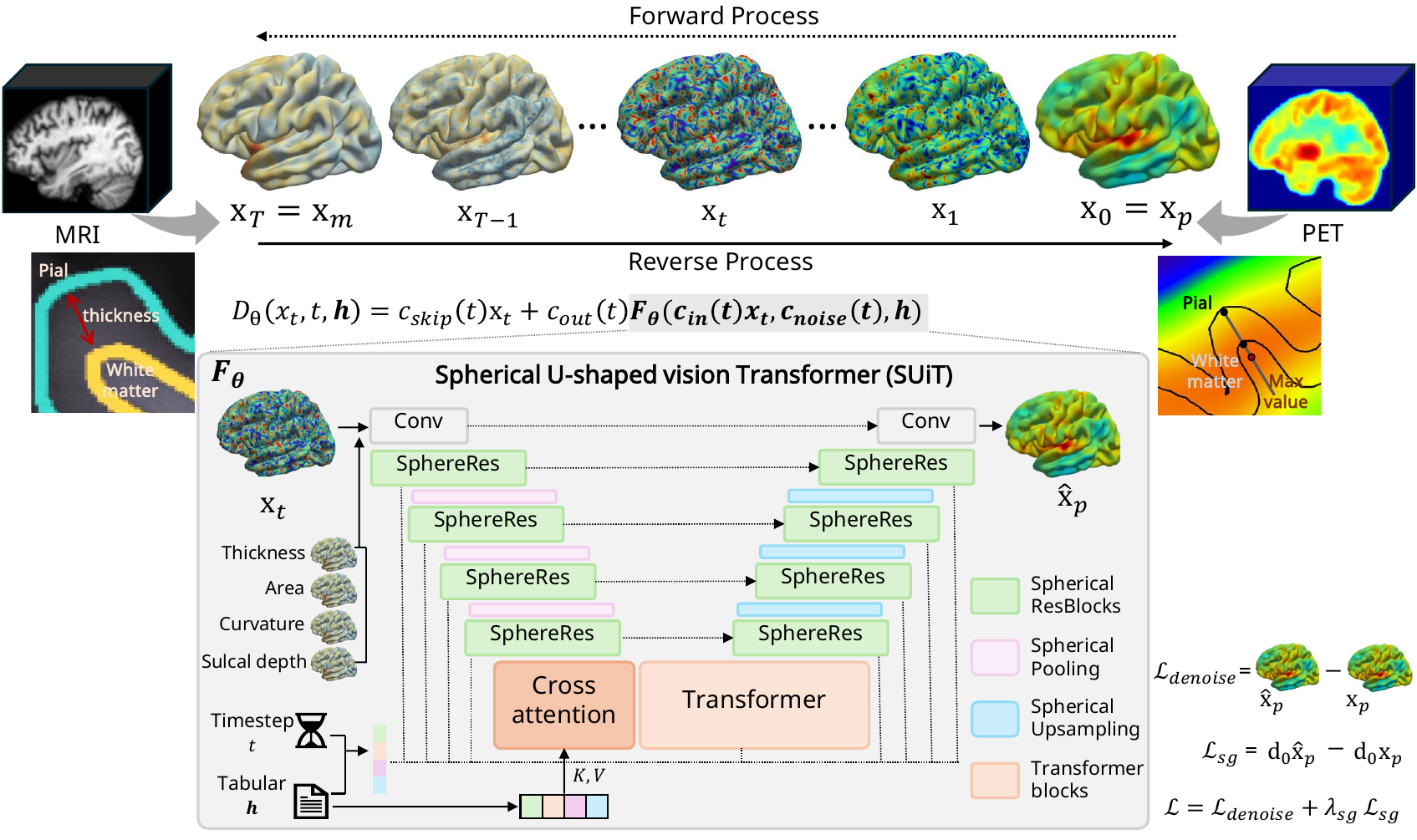}
    \caption{Overview of DB-SUiT, a surface-based conditional diffusion bridge framework. The cortical thickness map $\mathbf{x}_m$, derived from structural MRI, serves as the source modality, while the target FDG-PET surface $\mathbf{x}_p$ is obtained by sampling PET intensities along the pial–white matter trajectories and assigning the maximum value to each cortical vertex. The diffusion bridge denoising network $F\theta$ is implemented with a Spherical U-shaped Vision Transformer (SUiT), which combines topology-preserving spherical convolutions and global self-attention to model both local and distributed cortical patterns. Subject-level tabular features $\mathbf{h}$ are incorporated through cross-attention at the bottleneck. Training optimizes a weighted denoising loss $\mathcal{L}_{\text{denoise}}$ together with a surface gradient regularization term $\mathcal{L}_{\text{sg}}$.}
    \label{fig:method}
    \vspace{-1em}
\end{figure*}

\subsection{Diffusion Bridge for Surface-based Translation}
Given a paired sample $(\mathbf{x}_p, \mathbf{x}_m)$, unlike conventional diffusion models~\cite{ho2020denoising} that generate $\mathbf{x}_p$ from unstructured noise conditioned on $\mathbf{x}_m$, we directly learn their inherent correlation by formulating the task as a diffusion bridge process~\cite{zhou24ddbm}. It defines a stochastic trajectory, formulated as an SDE, that interpolates directly between these two endpoints:
\begin{equation}
\begin{split}
    d \mathbf{x}_t &= \mathbf{f}(\mathbf{x}_t, t)dt + g(t)^2 \mathbf{h}(\mathbf{x}_t,t,\mathbf{x}_T,T)dt + g(t)d\mathbf{w}_t, \\
    \mathbf{x}_0 &= \mathbf{x}_p, \ \mathbf{x}_T = \mathbf{x}_m,
\end{split}
\end{equation}
where $\mathbf{f}: \mathbb{R}^d \times [0, T] \rightarrow \mathbb{R}^d$ is vector-valued drift function which we set to be $\mathbf{f}(\mathbf{x}_t,t)=(d\log\alpha_t/dt)\mathbf{x}_t$, $g: [0, T] \rightarrow \mathbb{R}$ is a scalar-valued diffusion coefficient for $g(t)^2=d\sigma^2_t/dt-2\sigma_t^2(d\log\alpha_t/dt)$, $\mathbf{h}(\mathbf{x}_t,t,\mathbf{x}_T,T) = \nabla_{\mathbf{x}_t} \log p(\mathbf{x}_T \mid \mathbf{x}_t)$ is the gradient of the log transition kernel from current state $t$ to the endpoint $T$, and $\mathbf{w}_t$ is a standard Wiener process (Brownian motion). This forward process produces intermediate states $\mathbf{x}_t$ for $t \in [0, T]$, which smoothly transition from $\mathbf{x}_p$ to $\mathbf{x}_m$
via Doob's $h$-transform~\cite{doob1984classical}, yielding the transition kernel $p(\mathbf{x}_t \mid \mathbf{x}_p, \mathbf{x}_m) = \mathcal{N}(\mathbf{x}_t;\, a_t \mathbf{x}_m + b_t \mathbf{x}_p,\, \hat\sigma_t^2 \mathbf{I})$ with:
\begin{equation}
\begin{split}
    a_t = \frac{\mathrm{SNR}_T}{\mathrm{SNR}_t} \cdot \frac{\alpha_t}{\alpha_T}, & \qquad
    b_t = \alpha_t \!\left(1 - \frac{\mathrm{SNR}_T}{\mathrm{SNR}_t}\right), \\
    \hat\sigma_t^2 &= \sigma_t^2 \!\left(1 - \frac{\mathrm{SNR}_T}{\mathrm{SNR}_t}\right)\,,
\end{split}
\end{equation}
where $\alpha_t$ and $\sigma_t$ are signal and noise schedules, and $\mathrm{SNR}_t = \alpha_t^2 / \sigma_t^2$ is the signal-to-noise ratio at time $t$.
We adopt the variance-preserving (VP) formulation, where $\alpha_t = \exp{(-\tfrac{1}{4}\beta_d t^2 - \tfrac{1}{2}\beta_{\min} t)}$, $\sigma_t^2 = 1 - \alpha_t^2$, and $\beta_d$ and $\beta_{\min}$ are constant parameters.
Therefore, at $t{=}0$, the process recovers $\mathbf{x}_p$, and at $t{=}T$, it recovers $\mathbf{x}_m$, with controlled stochasticity at intermediate timesteps.
During training, we leverage a denoiser model $D_\theta$ that learns to denoise $\mathbf{x}_t$ back to $\mathbf{x}_p$, parameterized following the preconditioning framework of EDM~\cite{karras2022elucidating}:
\begin{equation}
    D_\theta(\mathbf{x}_t, t) = c_{\text{skip}}(t) \cdot \mathbf{x}_t + c_{\text{out}}(t) \cdot F_\theta\!\left(c_{\text{in}}(t) \cdot \mathbf{x}_t,\, c_{\text{noise}}(t)\right),
\end{equation}
where $F_\theta$ is a neural network, $c_{\text{skip}}$, $c_{\text{out}}$, $c_{\text{in}}$, are $c_{\text{noise}}$ are time-dependent scaling functions, derived to be:
\begin{align}
c_{\text{in}}(t) &= \frac{1}{\sqrt{a_t^2 \sigma_T^2 + b_t^2 \sigma_0^2 + 2 a_t b_t \sigma_{0T} + c_t}}, \\
c_{\text{out}}(t) &= \sqrt{a_t^2 (\sigma_T^2 \sigma_0^2 - \sigma_{0T}^2) + \sigma_0^2 c_t} \ast c_{\text{in}}(t), \\
c_{\text{skip}}(t) &= \left( b_t \sigma_0^2 + a_t \sigma_{0T} \right) \ast c_{\text{in}}^2(t), \\
c_{\text{noise}}(t) &= \frac{1}{4} \log(t) ,
\end{align}
in which $c_{\text{in}}$ rescales the noisy input to unit variance to stabilize network conditioning across noise levels; $c_{\text{out}}$ scales the network prediction to unit variance; $c_{\text{skip}}$ controls how much of the noisy input is carried directly to the prediction (dominant at low noise, where $x_t\approx x_0$) versus relying on the network (dominant at high noise); and $c_{\text{noise}}$ is a monotonic transform of the timestep/noise level used as the network's conditioning input. Together, they keep the effective training target and inputs well-scaled across the diffusion bridge process, stabilising model optimisation. 


\subsection{Deterministic Inference}
A notable characteristic of DB-SUiT is that the PET synthesis is deterministic at inference. Although stochastic noise perturbations are employed during training for multi-noise-level supervision and improved optimization, consistent with standard diffusion-model optimization, inference is performed by solving the reverse-time probability flow ODE using a second-order Heun solver. 
The probability flow ODE has the same marginal distributions as the diffusion bridge SDE, which is deterministic given the initial condition.
Since the process is initialized from the subject's MRI $\mathbf{x}_m$ rather than random noise, the MRI-to-PET generation is fully deterministic. Empirically, repeated inference under different random seeds yielded identical outputs. Consequently, DB-SUiT does not introduce sampling-related variability during deployment, alleviating concerns regarding uncertainty arising from stochastic image generation.

\subsection{Spherical U-shaped vision Transformer (SUiT)}
The denoising network $F_\theta$ should capture both local structural patterns, e.g., regional thinning, and long-range functional dependencies manifested as distributed hypometabolism across distant cortical regions.
However, existing surface models address these aspects in isolation: spherical U-Nets~\cite{zhao2019spherical} provide strong local, multi-scale representations but limited global context, whereas surface vision transformers~\cite{dahan2022surface} prioritize global interactions. We therefore
propose Spherical U-shaped vision Transformers (SUiT) that combine topology-preserving spherical convolutions with global self-attention to jointly model local and distributed cortical signals.
SUiT follows an encoder-bottleneck-decoder structure with multimodal conditioning (Fig.~\ref{fig:method}). The encoder consists of spherical ResNets followed by spherical pooling. As standard convolutions assume regular grid structures and cannot be directly applied to the irregular connectivity of surface meshes, we adopt spherical convolutions defined on the icosahedral mesh, where each vertex has a 1-ring neighborhood of 5 or 6 adjacent vertices. 
The convolutions preserve mesh topology by aggregating information only from geometrically adjacent vertices on the spherical manifold. 
Each ResNet block applies two spherical convolutions with group normalization and SiLU activation.
Feature maps are downsampled via spherical pooling over the 1-ring neighborhood from ico-$(k)$ to ico-$(k{-}1)$, 
reducing the vertex count by approximately a factor of~4. Each vertex at the coarser level corresponds to a well-defined sub-mesh at the finer level, 
providing natural multi-scale processing. 
Additional surface-derived MRI features, e.g., curvature, surface area, and sulcal depth, can be flexibly added as extra input channels to the encoder.
At the encoder's lowest resolution, we insert Transformer blocks to apply global self-attention across all remaining vertices, modeling long-range interactions across the entire cortex. 
The decoder mirrors the encoder with spherical upsampling, i.e., transposed spherical convolution, and skip connections 
from the corresponding encoder stage. 

To integrate complementary global context, SUiT is further augmented by the subject-level tabular features $\mathbf{h} \in \mathbb{R}^{38}$ via a dual conditioning scheme. 
The feature vector $\mathbf{h}$ is first mapped through an MLP to the time-embedding dimension and added to the timestep embedding, conditioning both the spherical ResNet blocks and the bottleneck Transformers via adaptive layer normalization (adaLN)~\cite{peebles2023scalable} to provide a global context signal.
Further, $\mathbf{h}$ is projected into $K$ learnable tokens aligned with the bottleneck feature dimension. These tokens are injected through multi-head cross-attention, where surface features act as queries and the conditional tokens as keys and values, enabling each vertex to attend to global subject information. This dual conditioning approach effectively integrates subject-level features at multiple levels of abstraction.

\subsection{Training Objective}
Our training objective combines a weighted denoising loss with a surface gradient regularization term:
$\mathcal{L} = \mathcal{L}_{\text{denoise}} + \lambda_{\text{sg}} \, \mathcal{L}_{\text{sg}}$.
Here, $\mathcal{L}_{\text{denoise}}$ is the weighted error between the denoised output $D_\theta$ and the ground-truth PET surface $\mathbf{x}_p$:
\begin{equation}
    \mathcal{L}_{\text{denoise}} = \mathbb{E}_{t,\, \mathbf{x}_p,\, \mathbf{x}_m,\, \boldsymbol{\epsilon}, \, \mathbf{h}} \left[ w(t) \left\| D_\theta(\mathbf{x}_t, t, \mathbf{h}) - \mathbf{x}_p \right\|^2 \right],
\end{equation}
where $w(t)$ follows the EDM weighting schedule~\cite{karras2022elucidating} that emphasizes noise levels where the denoising signal is most informative. 
We further introduce a surface gradient loss $\mathcal{L}_{\text{sg}}$ which leverages the discrete exterior derivative $d_0: \Omega^0(\mathcal{V}) \to \Omega^1(\mathcal{E})$ to map scalar vertex fields to edge-wise differences along mesh edges $\mathcal{E}$. This regularization term penalizes errors in the discrete metabolic gradient field:
\begin{equation*}
\mathcal{L}_{\text{sg}} = \frac{1}{|\mathcal{E}|} \sum_{e_{ij} \in \mathcal{E}} \left| (d_0\hat{\mathbf{x}}_p)(e_{ij}) - (d_0\mathbf{x}_p)(e_{ij}) \right|
\end{equation*}
where $\hat{\mathbf{x}}_p$, $\mathbf{x}_p$ are the predicted and ground-truth PET values, and $d_0\mathbf{x}_p(e_{ij}) = \mathbf{x}_p(v_j) - \mathbf{x}_p(v_i)$ represents the metabolic gradient along edge $e_{ij}$ connecting vertices $v_i$ and $v_j$. Modulated by a constant $\lambda_{sg}$, $\mathcal{L}_{\text{sg}}$ guides the model to faithfully reconstruct localized metabolic transition patterns on the cortex.

\begin{table*}[t]
\centering
\caption{Quantitative comparisons evaluated on both ADNI and in-house datasets. $^{\dagger}$For SiM2P, we projected their volumetric synthesis outputs onto the cortical surface using the same pipeline, as a fair comparison with the volumetric-based method.}
\label{tab:quantitative_compare}
\setlength{\tabcolsep}{8pt}
\begin{tabular}{lcccccc}
\toprule
Dataset & \multicolumn{3}{c}{ADNI} & \multicolumn{3}{c}{In-house} \\
\cmidrule(lr){2-4}\cmidrule(lr){5-7}
Method & MAE $\downarrow$ & PSNR $\uparrow$ & PCC $\uparrow$ & MAE $\downarrow$ & PSNR $\uparrow$ & PCC $\uparrow$ \\
\midrule
MLP     & \mystat{0.0608}{0.0359} & \mystat{24.03}{3.95} & \mystat{0.8824}{0.0529} & \mystat{0.0642}{0.0355} & \mystat{23.33}{3.72} & \mystat{0.8142}{0.0773} \\
S-UNet\cite{zhao2019spherical}  & \mystat{0.0563}{0.0199} & \mystat{23.43}{2.01} & \mystat{0.8748}{0.0481} & \mystat{0.0550}{0.0231} & \mystat{24.13}{2.95} & \mystat{0.8232}{0.0611} \\
SiT\cite{dahan2022surface} & \mystat{0.0476}{0.0223} & \mystat{25.31}{2.85} & \mystat{0.8701}{0.0554}  & \mystat{0.0540}{0.0219} & \mystat{24.23}{2.89} & \mystat{0.8177}{0.0706} \\
MS-SiT\cite{dahan2024multiscale}    & \mystat{0.0484}{0.0230} & \mystat{25.22}{2.90} & \mystat{0.8686}{0.0554} & \mystat{0.0567}{0.0237} & \mystat{23.85}{2.93} & \mystat{0.8002}{0.0730} \\
Pix2Pix\cite{isola2017image} & \mystat{0.0491}{0.0200} & \mystat{24.69}{2.33} & \mystat{0.8371}{0.0505} & \mystat{0.0579}{0.0182} & \mystat{23.28}{2.19} & \mystat{0.7390}{0.0843} \\
SUiT & \mystat{0.0472}{0.0230} & \mystat{25.45}{2.97} & \mystat{0.8775}{0.0557} & \mystat{0.0548}{0.0231} & \mystat{24.17}{2.96} & \mystat{0.8234}{0.0634} \\\midrule
SiM2P$^{\dagger}$~\cite{li2025diffusion}   &     \mystat{0.0547}{0.0272}    &       \mystat{24.16}{3.15} &    \mystat{0.8353}{0.0586}   & \mystat{0.0568}{0.0249}     & \mystat{23.79}{3.07} & \mystat{0.7879}{0.0608} \\
BBDM (SUiT)   &     \mystat{0.0456}{0.0218}    &       \mystat{25.64}{2.79}   &  \mystat{\underline{0.8841}}{0.0480}  & \mystat{0.0480}{0.0154}     & \mystat{\underline{24.97}}{2.27} & \mystat{0.8485}{0.0364} \\
DDBM (S-Unet)    & \mystat{0.0480}{0.0221} & \mystat{25.13}{2.70} & \mystat{0.8645}{0.0483} & \mystat{\underline{0.0479}}{0.0160} & \mystat{\underline{24.97}}{2.24} & \mystat{\underline{0.8500}}{0.0372} \\
DDBM (SiT)               & \mystat{0.0494}{0.0229} & \mystat{24.84}{2.66} & \mystat{0.8505}{0.0508} & \mystat{0.0510}{0.0177} & \mystat{24.40}{2.27} & \mystat{0.8184}{0.0416} \\
DDBM (MS-SiT) & \mystat{\underline{0.0454}}{0.0220} & \mystat{\underline{25.65}}{2.78} & \mystat{0.8803}{0.0497} & \mystat{0.0497}{0.0175} & \mystat{24.66}{2.32} & \mystat{0.8317}{0.0406} \\
\midrule
\textbf{DB-SUiT (Ours)}            & \mystat{\textbf{0.0430}}{0.0216} & \mystat{\textbf{26.17}}{2.84} & \mystat{\textbf{0.8969}}{0.0491} & 
\mystat{\textbf{0.0438}}{0.0141} & \mystat{\textbf{25.69}}{2.21} & \mystat{\textbf{0.8676}}{0.0319} \\
\bottomrule
\end{tabular}
\vspace{-1em}
\end{table*}

\section{Experiment Setup}

\paragraph{\textbf{Data and Preprocessing}} 
We use paired FDG-PET and T1-weighted MRI scans from two datasets: Alzheimer’s disease neuroimaging initiative (ADNI)~\cite{adni}, including cognitively normal (CN, $n=379$), subjects with mild cognitive impairment (MCI, $n=611$), and Alzheimer’s disease (AD, $n=257$); a single-site in-house clinical dataset from the TUM (Technical University of Munich, Germany) University hospital containing two types of dementia, with 143 CN, 110 AD, and 57 frontotemporal dementia (FTD) samples. 
The MRI surface and subcortical measures, in total 36, were obtained from FreeSurfer v7.2~\cite{fischl2012freesurfer}, including ventricular, cerebellar, subcortical gray-matter, white-matter, corpus-callosum, and global intracranial/brain-volume measures. 
The PET surface projection followed the established pipeline~\cite{marcoux2018automated}: the PET volume was first registered to the subject’s MRI volume, normalized, and then projected onto the cortical surface (reconstructed by FreeSurfer~\cite{fischl2012freesurfer} from MRI) by sampling intensities along the line between corresponding pial and white matter vertices and assigning the maximum value to each vertex. Based on Minoshima et al.~\cite{minoshima202218f}, the sampling line was extended 5 mm into the white matter to account for registration errors. 

\paragraph{\textbf{Implementation Details}} 
SUiT employs a base dimension $d = 128$, channel multipliers $(1, 2, 4, 8)$ across 4 resolutions, 4 bottleneck Transformer blocks with 8 attention heads, and $K=16$. We select $\lambda_{\text{sg}} = 0.1$ after an exhaustive hyperparameter search on the validation set, and a sampling step of 50.
The model is trained using Adam optimizer with a batch size of~8, a learning rate of $10^{-4}$, and no weight decay on a single NVIDIA A100 GPU. 

\paragraph{\textbf{Baselines}}
The baselines include non-diffusion-based translation models MLP, spherical U-Net (S-UNet)~\cite{zhao2019spherical}, surface vision Transformer (SiT)~\cite{dahan2022surface}, multiscale SiT (MS-SiT)~\cite{dahan2024multiscale}, our SUiT backbone, an established GAN-based image-to-image translation approach Pix2Pix~\cite{isola2017image}, and diffusion model-based approaches BBDM~\cite{bbdm} and DDBM~\cite{zhou24ddbm}, implemented with aforementioned backbones.
In addition, we evaluated the volumetric baseline SiM2P~\cite{li2025diffusion} by projecting its volumetric synthesis outputs onto the cortical surface using the identical processing pipeline and subsequently computing all evaluation metrics in the surface domain. This ensured a fair and consistent comparison between the proposed surface-based framework and a representative volume-based translation method.

\paragraph{\textbf{Evaluation Setup}} 
All methods are evaluated both quantitatively and qualitatively. Quantitative metrics include mean absolute error (MAE), peak signal-to-noise ratio (PSNR), and Pearson correlation coefficient (PCC) between real and synthesized PET surfaces. 
To assess whether the synthesized PET surfaces preserve diagnostically relevant information, we evaluated automated classification of differential diagnosis of dementia (CN vs. AD vs. bvFTD) on the in-house data with 5-fold cross-validation (both the DB-SUiT model and the downstream classifier trained and evaluated independently in each fold), as well as two early-stage disease progression tasks, CN vs. MCI and MCI vs. AD, on the ADNI dataset. Qualitative results are presented as synthesized surfaces alongside error and Z-score maps, computed on a per-vertex basis relative to a healthy reference cohort of 40 subjects negative for amyloid-beta, tau or neurodegeneration (A-/T-/N-). The healthy reference cohort comprised subjects from a held-out ADNI test set.

\paragraph{\textbf{Clinical Validation}} 
To evaluate the clinical utility of DB-SUiT synthetic PET surfaces in dementia diagnosis, we conducted a blinded reader study on an external in-house set of 62 subjects (cognitively normal (CN) controls, n = 22; Alzheimer’s disease (AD), n = 19; behavioral-variant frontotemporal dementia (bvFTD), n = 21). 
Synthetic PET surfaces were generated by directly applying the ADNI-trained DB-SUiT model without any fine-tuning, evaluating its ability to generalize to both an unseen cohort and an unseen pathology.
The study involved a board-certified senior nuclear medicine physician and a board-certified senior neuroradiologist, each with 14 years of clinical and subspecialty experience. The nuclear medicine physician independently assessed real and synthetic PET surfaces together with corresponding Z-score maps, whereas the neuroradiologist assessed real MRI, with age and sex provided in all settings. Both readers were blinded to the ground-truth diagnoses and label distribution. For each case, they performed a two-stage assessment: first, determining whether dementia was present and reporting diagnostic confidence (low, moderate, or high); and second, for cases judged as dementia, classifying the pattern as AD or bvFTD with a corresponding confidence level.

As diagnostic confidence is directly relevant to clinical utility, we additionally computed confidence-weighted accuracy. For each case $i$, the correctness indicator $y_i$ ($1$ if correct, $0$ otherwise) was multiplied by a confidence weight $w_i$ (low = 1.0, moderate = 2.0, high = 3.0), and weighted accuracy was calculated as $\sum_i w_i y_i / \sum_i w_i$. This metric rewards correct high-confidence decisions while downweighting uncertain assessments, providing a clinically informative complement to plain accuracy.

\begin{figure*}[t]
    \centering
    \includegraphics[width=0.9\linewidth]{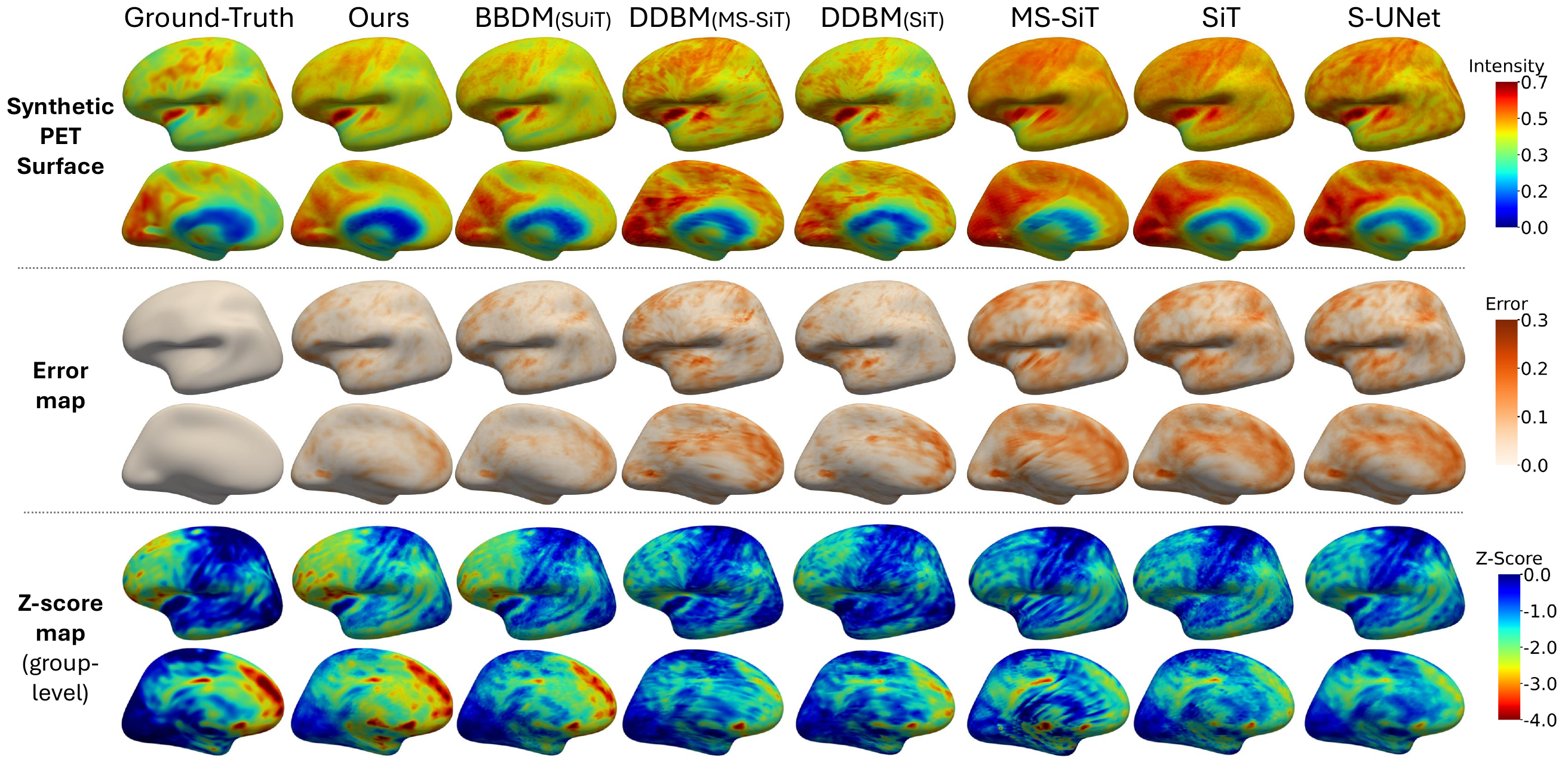}
    \caption{Qualitative results of generated PET surfaces from our proposed method and different baselines compared to the ground-truth PET surface, with error maps and group-level z-score maps on both lateral and medial views.}
    \label{fig:qualitative_compare}
\end{figure*}

\section{Results}

\subsection{Quantitative Evaluation} 
Tab.~\ref{tab:quantitative_compare} presents a comprehensive quantitative evaluation of DB-SUiT against baselines on both ADNI and in-house datasets. We trained separate models exclusively for each cohort.
The results reveal three key findings. First, the proposed SUiT backbone yields competitive performance even in the deterministic (non-diffusion) setting and further improves results when integrated into diffusion models, highlighting its strong capacity for modeling cortical surface topology and long-range dependencies. Second, diffusion-based methods generally outperform their non-diffusion counterparts, confirming the advantage of probabilistic bridge modeling for cross-modal synthesis. 
This advantage is particularly relevant for MRI-to-PET translation, where the mapping can be inherently heterogeneous, especially at early disease stages; direct regression methods may therefore collapse toward the conditional mean and produce over-smoothed estimates, whereas diffusion models learn a richer conditional distribution through multi-level denoising supervision and can better preserve subtle metabolic patterns while still allowing deterministic inference via the probability-flow ODE.
Finally, our proposed framework DB-SUiT consistently surpasses all other configurations across datasets, attaining the best performance with an MAE of 0.0430 and PSNR of 26.17 dB on ADNI, and an MAE of 0.0438 and PSNR of 25.69 dB on the in-house dataset, which reduces MAE by 5.3\% on ADNI and 8.6\% on the in-house dataset compared with the strongest competing method, highlighting its synergistic power to achieve high-fidelity and robust cross-modal translation.
Notably, compared with the volumetric SiM2P~\cite{li2025diffusion} baseline projected onto the cortical surface, DB-SUiT yields substantially lower errors and higher correlations, with MAE reductions of 21.4\% on ADNI and 22.9\% on the in-house dataset, supporting the benefit of direct surface-domain modeling.

\subsection{Region-wise Reconstruction Fidelity Across Lobes}

To further assess whether DB-SUiT preserves PET reconstruction fidelity across anatomically distinct cortical regions, we performed a lobe-wise MAE analysis using the Desikan-Killiany atlas~\cite{DESIKAN2006968}. Results in Table~\ref{tab:lobe_mae} show that DB-SUiT achieved the lowest reconstruction error across all six cortical lobes on both ADNI and the in-house dataset, including frontal, parietal, temporal, occipital, cingulate, and insular regions. On ADNI, DB-SUiT reduced MAE to 0.0429, 0.0464, 0.0389, 0.0545, 0.0406, and 0.0424 in these regions, respectively. The same trend was observed on the in-house cohort, with corresponding MAEs of 0.0411, 0.0484, 0.0399, 0.0516, 0.0436, and 0.0468. Compared with both deterministic surface models and alternative bridge-based baselines, DB-SUiT showed consistent improvements rather than gains limited to a single anatomical region. These results suggest that the proposed diffusion-bridge formulation improves surface PET synthesis in a spatially robust manner, supporting accurate reconstruction of distributed cortical metabolic patterns relevant to dementia characterization.

\begin{table*}[t]
\centering
\setlength{\tabcolsep}{6pt}
\caption{Region-wise reconstruction error (MAE~$\downarrow$, mean$\pm$std) on the ADNI and in-house datasets. Cortical vertices are parcellated with the Desikan-Killiany atlas and grouped into lobes; the medial wall is excluded.}
\label{tab:lobe_mae}
\begin{tabular}{lcccccc}
\toprule
Lobe & \multicolumn{2}{c}{\textbf{Frontal}} & \multicolumn{2}{c}{\textbf{Parietal}} & \multicolumn{2}{c}{\textbf{Temporal}} \\
\cmidrule(lr){2-3}\cmidrule(lr){4-5}\cmidrule(lr){6-7}
Method & ADNI & In-house & ADNI & In-house & ADNI & In-house \\
\midrule
MLP & \mystat{0.0631}{0.0419} & \mystat{0.0656}{0.0396} & \mystat{0.0654}{0.0410} & \mystat{0.0662}{0.0402} & \mystat{0.0550}{0.0346} & \mystat{0.0597}{0.0386} \\
S-UNet\cite{zhao2019spherical} & \mystat{0.0467}{0.0268} & \mystat{0.0502}{0.0230} & \mystat{0.0508}{0.0277} & \mystat{0.0632}{0.0305} & \mystat{0.0420}{0.0219} & \mystat{0.0507}{0.0240} \\
SiT\cite{dahan2022surface} & \mystat{0.0466}{0.0254} & \mystat{0.0501}{0.0223} & \mystat{0.0530}{0.0281} & \mystat{0.0633}{0.0303} & \mystat{0.0436}{0.0234} & \mystat{0.0501}{0.0236} \\
MS-SiT\cite{dahan2024multiscale} & \mystat{0.0468}{0.0261} & \mystat{0.0517}{0.0244} & \mystat{0.0533}{0.0287} & \mystat{0.0661}{0.0327} & \mystat{0.0442}{0.0245} & \mystat{0.0524}{0.0266} \\
SUiT & \mystat{0.0459}{0.0261} & \mystat{0.0506}{0.0236} & \mystat{0.0525}{0.0285} & \mystat{0.0643}{0.0316} & \mystat{0.0433}{0.0240} & \mystat{0.0511}{0.0249} \\
BBDM (SUiT) & \mystat{0.0456}{0.0260} & \mystat{0.0484}{0.0183} & \mystat{\underline{0.0493}}{0.0261} & \mystat{0.0518}{0.0215} & \mystat{0.0412}{0.0209} & \mystat{0.0457}{0.0155} \\
DDBM (S-Unet) & \mystat{0.0480}{0.0263} & \mystat{\underline{0.0474}}{0.0165} & \mystat{0.0519}{0.0271} & \mystat{\underline{0.0510}}{0.0238} & \mystat{0.0435}{0.0205} & \mystat{\underline{0.0449}}{0.0169} \\
DDBM (SiT) & \mystat{0.0497}{0.0272} & \mystat{0.0494}{0.0181} & \mystat{0.0536}{0.0280} & \mystat{0.0548}{0.0253} & \mystat{0.0442}{0.0222} & \mystat{0.0488}{0.0190} \\
DDBM (MS-SiT) & \mystat{\underline{0.0453}}{0.0259} & \mystat{0.0476}{0.0177} & \mystat{0.0498}{0.0262} & \mystat{0.0557}{0.0246} & \mystat{\underline{0.0410}}{0.0220} & \mystat{0.0458}{0.0172} \\
\addlinespace[2pt]\midrule\addlinespace[2pt]
\textbf{DB-SUiT (Ours)} & \mystat{\textbf{0.0429}}{0.0258} & \mystat{\textbf{0.0411}}{0.0151} & \mystat{\textbf{0.0464}}{0.0256} & \mystat{\textbf{0.0484}}{0.0214} & \mystat{\textbf{0.0389}}{0.0211} & \mystat{\textbf{0.0399}}{0.0160} \\
\bottomrule
\end{tabular}

\vspace{1.5mm}

\begin{tabular}{lcccccc}
\toprule
Lobe & \multicolumn{2}{c}{\textbf{Occipital}} & \multicolumn{2}{c}{\textbf{Cingulate}} & \multicolumn{2}{c}{\textbf{Insula}} \\
\cmidrule(lr){2-3}\cmidrule(lr){4-5}\cmidrule(lr){6-7}
Method & ADNI & In-house & ADNI & In-house & ADNI & In-house \\
\midrule
MLP & \mystat{0.0720}{0.0498} & \mystat{0.0707}{0.0483} & \mystat{0.0591}{0.0368} & \mystat{0.0684}{0.0440} & \mystat{0.0600}{0.0373} & \mystat{0.0687}{0.0447} \\
S-UNet\cite{zhao2019spherical} & \mystat{0.0571}{0.0365} & \mystat{0.0596}{0.0348} & \mystat{0.0453}{0.0236} & \mystat{0.0581}{0.0296} & \mystat{0.0442}{0.0232} & \mystat{0.0534}{0.0241} \\
SiT\cite{dahan2022surface} & \mystat{0.0567}{0.0348} & \mystat{0.0599}{0.0361} & \mystat{0.0477}{0.0240} & \mystat{0.0585}{0.0284} & \mystat{0.0451}{0.0218} & \mystat{0.0534}{0.0249} \\
MS-SiT\cite{dahan2024multiscale} & \mystat{0.0579}{0.0352} & \mystat{0.0616}{0.0342} & \mystat{0.0487}{0.0246} & \mystat{0.0609}{0.0286} & \mystat{0.0449}{0.0223} & \mystat{0.0566}{0.0248} \\
SUiT & \mystat{0.0565}{0.0352} & \mystat{0.0600}{0.0353} & \mystat{0.0470}{0.0250} & \mystat{0.0590}{0.0307} & \mystat{0.0437}{0.0225} & \mystat{0.0535}{0.0244} \\
BBDM (SUiT) & \mystat{0.0577}{0.0353} & \mystat{0.0554}{0.0224} & \mystat{\underline{0.0442}}{0.0237} & \mystat{\underline{0.0497}}{0.0219} & \mystat{0.0435}{0.0228} & \mystat{\underline{0.0484}}{0.0217} \\
DDBM (S-Unet) & \mystat{0.0599}{0.0360} & \mystat{\underline{0.0537}}{0.0229} & \mystat{0.0463}{0.0234} & \mystat{0.0521}{0.0205} & \mystat{0.0460}{0.0223} & \mystat{0.0504}{0.0206} \\
DDBM (SiT) & \mystat{0.0614}{0.0377} & \mystat{0.0596}{0.0260} & \mystat{0.0485}{0.0240} & \mystat{0.0539}{0.0226} & \mystat{0.0478}{0.0234} & \mystat{0.0515}{0.0198} \\
DDBM (MS-SiT) & \mystat{\underline{0.0565}}{0.0351} & \mystat{0.0564}{0.0275} & \mystat{0.0443}{0.0231} & \mystat{0.0533}{0.0215} & \mystat{\underline{0.0427}}{0.0224} & \mystat{0.0499}{0.0197} \\
\addlinespace[2pt]\midrule\addlinespace[2pt]
\textbf{DB-SUiT (Ours)} & \mystat{\textbf{0.0545}}{0.0339} & \mystat{\textbf{0.0516}}{0.0216} & \mystat{\textbf{0.0406}}{0.0234} & \mystat{\textbf{0.0436}}{0.0156} & \mystat{\textbf{0.0424}}{0.0235} & \mystat{\textbf{0.0468}}{0.0218} \\
\bottomrule
\end{tabular}
\vspace{-1em}
\end{table*}

\subsection{Qualitative Fidelity and Pathological Interpretability} 
Beyond numerical metrics, we provide qualitative comparisons for a representative FTD subject in Fig.~\ref{fig:qualitative_compare} to illustrate the model's ability to reconstruct metabolic patterns with pathological details. DB-SUiT produces PET surfaces that most closely match the ground truth in both global intensity and regional patterns without over-smoothing. 
Using SUiT in BBDM also reduces reconstruction errors, further highlighting its architectural advantage in high-fidelity surface generation.
In contrast, other baselines exhibit either blurred metabolic patterns or regional biases.
Fig.~\ref{fig:teaser} also shows our surface-based synthesis preserves more accurate pathological details than the volume-based diffusion bridge method~\cite{li2025diffusion}, underscoring the necessity of a geometrically grounded framework.
To further assess pathological accuracy at the group level, we computed averaged z-score maps for all FTD test subjects relative to healthy controls. DB-SUiT successfully captures the hallmark signature of FTD, characterized by distinct frontal and temporal hypometabolism, mirroring the ground-truth distribution with high precision.



\begin{table*}[t]
\centering
\caption{Classification results for differential dementia diagnosis on the in-house dataset. 
Results are reported as mean $\pm$ standard deviation across folds.}
\label{tab:classification}
\setlength{\tabcolsep}{14pt}
\begin{tabular}{lllcc}
\toprule
\multirow{2}{*}{\textbf{Input Modality}} & \multirow{2}{*}{\textbf{Format}} & \multirow{2}{*}{\textbf{Feature / model}} 
& \multicolumn{2}{c}{\textbf{CN vs. AD vs. FTD}} \\
\cmidrule(lr){4-5}
 & & & \textbf{BACC (\%)$\uparrow$} & \textbf{F1-score (\%)$\uparrow$} \\
\midrule

\multirow{4}{*}{MRI}
& volume  & intensity                  & \mystat{65.99}{6.58} & \mystat{62.58}{7.81} \\
& surface & thickness                  & \mystat{66.56}{3.34} & \mystat{69.41}{3.73} \\
& surface & thickness + tabular         & \mystat{67.73}{4.66} & \mystat{69.93}{5.37} \\
& surface & thickness + area + curv. + sulc. + tabular 
                                      & \mystat{65.02}{1.16} & \mystat{66.42}{1.63} \\

\midrule
\multirow{2}{*}{Real PET}
& volume  & intensity                  & \mystat{69.47}{5.05} & \mystat{69.53}{5.84} \\
& surface & intensity                  & \mystat{\textbf{79.34}}{5.25} & \mystat{\textbf{82.49}}{4.96} \\

\midrule
\multirow{3}{*}{Synthetic PET}
& surface & intensity / DDBM$_{\mathrm{SiT}}$       & \mystat{65.78}{5.50} & \mystat{67.70}{4.22} \\
& surface & intensity / DDBM$_{\mathrm{SUnet}}$     & \mystat{69.51}{7.72} & \mystat{70.08}{7.54} \\
& surface & intensity / DB-SUiT (ours)              & \mystat{\underline{77.32}}{3.29} & \mystat{\underline{78.54}}{2.92} \\

\bottomrule
\end{tabular}
\vspace{-1em}
\end{table*}


\begin{table}[h!]
\centering
\caption{Classification results on the ADNI dataset. We report both BACC (\%) and F1-score (\%) on the two tasks (CN vs. MCI and MCI vs. AD), comparing different input modalities with our DB-SUiT synthesized PET (Syn PET).}
\label{tab:adni_classification}
\setlength{\tabcolsep}{2pt}
\begin{tabular}{lllcccc}
\toprule
& & & \multicolumn{2}{c}{\textbf{CN vs. MCI}} & \multicolumn{2}{c}{\textbf{MCI vs. AD}} \\
\cmidrule(lr){4-5} \cmidrule(lr){6-7}
\textbf{Modality} & \textbf{Format} & \textbf{Feature} & \textbf{BACC$\uparrow$} & \textbf{F1$\uparrow$} & \textbf{BACC$\uparrow$} & \textbf{F1$\uparrow$} \\
\midrule
MRI & surface & thickness & 59.28 & 59.17 & 69.25 & 66.73 \\
MRI & surface & thick.+tab. & 60.59 & 59.49 & 71.25 & 68.20 \\
MRI & surface & thick/area/curv/sulc+tab. & 60.39 & 60.48 & 62.98 & 63.40 \\
GT PET & surface & intensity & \pmb{65.51} & \pmb{65.01} & \pmb{73.80} & \pmb{73.52} \\
Syn PET & surface & intensity & \underline{64.69} & \underline{64.09} & \underline{72.16} & \underline{71.37} \\
\bottomrule
\end{tabular}
\vspace{-1em}
\end{table}

\subsection{Classification Results for the Differential Diagnosis of Dementia} 
To assess the diagnostic utility of synthesized PET surfaces, we performed a three-way differential diagnosis (CN vs. AD vs. FTD), comparing DB-SUiT to different input modalities: (1) Surface-based features from MRI (with or without tabular data, extra conditions surface area, curvature, sulcal depth), ground-truth (GT) PET, and synthetic (Syn) PET from our and baseline methods. A three-layer MLP was employed with early-stop to prevent over-fitting;
(2) Volumetric features from standard 3D MRI and PET images evaluated with 3D vision Transformers~\cite{singla2022multiple}.
Crucially, the classifier for Syn PET surfaces was trained exclusively on GT PET, providing a rigorous test of both their pathological fidelity and domain alignment.
As shown in Table~\ref{tab:classification}, among volumetric inputs, PET volumes outperform MRI (BACC: 69.5 vs. 66.0\%), confirming the higher disease sensitivity of metabolic imaging. 
Transitioning to the surface domain further improves performance, with GT PET surfaces achieving the best overall results (BACC: 79.3\%, F1: 82.5\%). This improvement is likely twofold: first, the surface-based representation provides a more compact, geometrically aligned encoding of metabolic activity that minimizes confounding effects of partial volume averaging; second, the reduced dimensionality of the surface feature space allows for more efficient parameterization of the classification network, which is particularly advantageous for limited clinical data.
The synthetic PET surfaces yield the second-best diagnostic accuracy (BACC: 77.3\%, F1: 78.5\%), representing a 14.2\% significant improvement in BACC over the best MRI configuration ($p$-value$<$0.05, Wilcoxon signed-rank test), and a 11.2\% gain over Syn PET from baseline methods, approaching the accuracy of GT PET. 
We also evaluated DB-SUiT on the larger ADNI cohort for two early-stage diagnostic tasks: CN vs. MCI and MCI vs. AD. As shown in Table~\ref{tab:adni_classification}, for MCI vs. AD, synthetic PET achieved 72.2\% BACC and 71.4\% F1, outperforming MRI thickness alone (69.3\%/66.7\%) and MRI thickness with tabular data (71.3\%/68.2\%), and closely approaching real PET (73.8\%/73.5\%). For the more challenging CN vs. MCI task, synthetic PET achieved 64.7\% BACC and 64.1\% F1, again improving over MRI thickness alone (59.3\%/59.2\%) and MRI thickness with tabular data (60.6\%/59.5\%), while remaining close to real PET (65.5\%/65.0\%). 
These results demonstrate that DB-SUiT successfully translates structural and tabular inputs into diagnostically informative metabolic signals, providing consistent gains over MRI and approaching the utility of real PET across both differential dementia diagnosis and early-stage disease settings.

\subsection{Clinical Reader Study Validation}

\begin{table}[t]
\centering
\caption{Clinical reader study results using real and synthetic (Syn) PET surface. Standard accuracy (ACC) and confidence-weighted (Weighted) ones are reported for dementia detection and three-class differential diagnosis. }
\label{tab:reader_real_syn_pet}
\setlength{\tabcolsep}{2pt}
\begin{tabular}{lcccccc}
\toprule
\multirow{2}{*}{\textbf{Task}} 
& \multicolumn{2}{c}{\textbf{Real PET}} 
& \multicolumn{2}{c}{\textbf{MRI}}
& \multicolumn{2}{c}{\textbf{Syn PET}} \\
\cmidrule(lr){2-3} \cmidrule(lr){4-5} \cmidrule(lr){6-7}
& ACC & Weighted & ACC & Weighted & ACC & Weighted \\
\midrule
\multirow{1}{*}{\textbf{CN vs. dementia}}
 & 96.77 & 97.11 & 80.65 & 85.81 & 91.94 & 93.64 \\
\multirow{1}{*}{\textbf{CN vs. AD vs. bvFTD}}
 & 95.16 & 97.50 & 75.81 & 83.10 & 85.48 & 88.46 \\
\bottomrule
\end{tabular}
\vspace{-1em}
\end{table}

To evaluate whether synthetic PET surfaces from DB-SUiT can effectively support clinical decision-making in dementia diagnosis, we conducted a blinded reader study, using the DB-SUiT model trained solely on the ADNI dataset and applying it directly to an external in-house set across three groups: CN, AD, and bvFTD. A board-certified senior nuclear medicine physician independently evaluated real and synthetic PET surfaces, while a board-certified senior neuroradiologist evaluated real MRI, with age and sex provided in all settings.


As shown in Table~\ref{tab:reader_real_syn_pet}, real PET achieved consistently high performance, reaching 96.77\% standard and 97.11\% confidence-weighted accuracy for dementia detection (CN vs. dementia) and 95.16\%/97.50\% for the three-class differential diagnosis (CN vs. AD vs. bvFTD), confirming that surface-projected PET retains the metabolic information clinicians rely on. Performance using MRI was markedly weaker (80.65\%/85.81\% and 75.81\%/83.10\%) due to its limited specificity of atrophy patterns. 
Synthetic PET surface improved over MRI by 11.29 and 9.67 standard-accuracy points on the two tasks (7.83 and 5.36 points after confidence weighting), corresponding to a 58\% and 40\% relative reduction in reader error, and closing 70\% and 50\% of the MRI-to-real-PET gap. 
Together, these results show that synthetic PET surfaces convey diagnostic information well beyond what MRI alone supports and approaches the real PET reference.

\begin{figure*}[t]
    \centering
    \vspace{-1em}
    \includegraphics[width=0.9\linewidth]{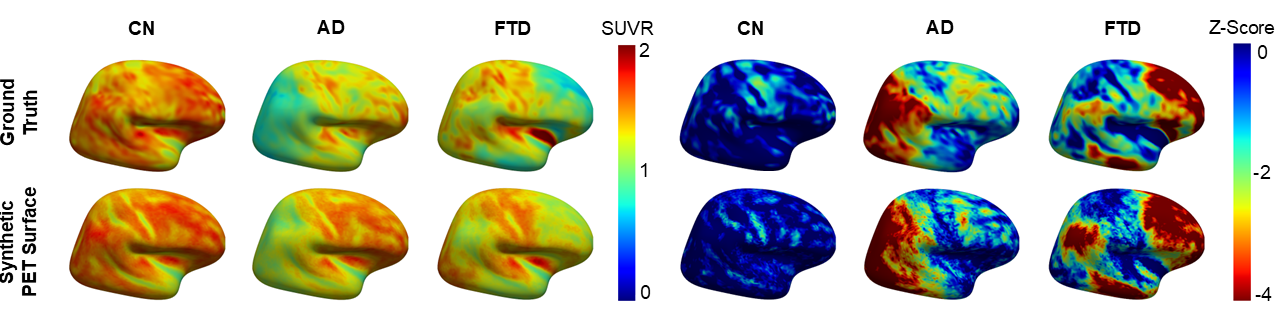}
    \vspace{-1em}
    \caption{
    Cross-cohort and cross-pathology generalization of DB-SUiT. The synthetic PET surfaces are generated from the in-house MRI input by the ADNI-pretrained DB-SUiT model, without fine-tuning or adaptation, compared to the corresponding ground-truth PET surfaces with Z-score maps for CN, AD, and FTD subjects. Despite being trained only on ADNI, the model generalizes well to the unseen in-house cohort and FTD pathology, which was not represented during training.
    }
    \label{fig:case_comparison}
    \vspace{-1em}
\end{figure*}

Notably, the synthetic PET surfaces used in the clinical reader study were generated by directly applying the ADNI-pretrained DB-SUiT model to the in-house MRI thickness maps, without any fine-tuning or adaptation. Despite never being trained on the in-house cohort or any FTD cases, the model successfully generalized to these unseen samples and produced synthetic PET surfaces of sufficient diagnostic quality for clinical evaluation (Figure~\ref{fig:case_comparison}). This strong performance demonstrates the excellent cross-cohort and cross-pathology generalizability of DB-SUiT.

\subsection{Ablation Study}
Table~\ref{tab:ablation} summarizes ablation studies on the ADNI validation set. Adding tabular data progressively improves the generation performance, with the inclusion of age, sex, and subcortical volumetric features achieving the best results.
This suggests that subject-level demographic and anatomical information provides complementary context beyond the cortical surface MRI features.
For the tabular conditioning strategy, both additive and cross-attention conditioning achieve comparable improvements, while their combination yields the best overall performance, with the former providing global modulation and the latter enabling more flexible feature interaction. 
The surface gradient loss further improves performance when $\lambda_{sg} \neq 0$, with the optimum observed at $\lambda_{sg} = 0.1$. Larger weights slightly degrade performance, suggesting that moderate gradient regularization helps preserve local surface consistency, whereas overly strong regularization may over-constrain the reconstruction. 
Finally, with cortical thickness as the source input for the diffusion bridge, adding additional surface conditions as extra input channels consistently improves performance, with the full set of area, curvature, and sulcal depth achieving the best overall results. 
These results validate the contribution of multimodal conditioning, balanced surface-gradient regularization, and surface geometry grounding.

\begin{table}[t]
\centering
\caption{Ablation studies on the ADNI validation set.}
\label{tab:ablation}
\setlength{\tabcolsep}{4pt}
\renewcommand{\arraystretch}{1.08}

\begin{tabularx}{\columnwidth}{
@{}>{\raggedright\arraybackslash}X
r@{\,${}\pm{}$\,}l
r@{\,${}\pm{}$\,}l@{}
}
\toprule
\textbf{Setting} 
& \multicolumn{2}{c}{\textbf{MAE $\downarrow$}} 
& \multicolumn{2}{c@{}}{\textbf{PSNR $\uparrow$}} \\
\midrule
\addlinespace[2pt]

\rowcolor{black!7}
\multicolumn{5}{@{}c@{}}{\textbf{Tabular input}} \\
\quad None              & 0.0446 & 0.024 & 25.96 & 3.08 \\
\quad Age + Sex           & 0.0443 & 0.023 & 25.98 & 3.02 \\
\quad Age + Sex + Volumes   & \textbf{0.0437} & 0.023 & \textbf{26.11} & 3.06 \\

\addlinespace[1pt]
\rowcolor{black!7}
\multicolumn{5}{@{}c@{}}{\textbf{Condition strategy}} \\
\quad Additive          & 0.0438 & 0.023 & 26.10 & 3.06 \\
\quad Cross-attention   & 0.0437 & 0.023 & 26.10 & 3.02 \\
\quad Both              & \textbf{0.0437} & 0.023 & \textbf{26.11} & 3.06 \\

\addlinespace[1pt]
\rowcolor{black!7}
\multicolumn{5}{@{}c@{}}{\textbf{\boldmath$\lambda_\mathrm{sg}$}} \\
\quad 0.0                 & 0.0441 & 0.023 & 26.03 & 3.04 \\
\quad 0.1               & \textbf{0.0437} & 0.023 & \textbf{26.11} & 3.06 \\
\quad 0.5               & 0.0439 & 0.023 & 26.07 & 3.04 \\
\quad 1.0               & 0.0441 & 0.023 & 26.02 & 3.04 \\


\addlinespace[1pt]
\rowcolor{black!7}
\multicolumn{5}{@{}c@{}}{\textbf{Extra condition}} \\
\quad None              & 0.0437 & 0.023 & 26.11 & 3.06 \\
\quad Area            & 0.0438 & 0.023 & 26.11 & 3.08 \\
\quad Area + Curvature       & 0.0435 & 0.023 & 26.22 & 3.12 \\
\quad Area + Curvature + Sulcal depth    & \textbf{0.0433} & 0.023 & \textbf{26.27} & 3.15 \\

\bottomrule
\end{tabularx}
\vspace{-1em}
\end{table}

\section{Discussion}

DB-SUiT introduces the first surface-based MRI-to-PET translation framework on the cortical manifold, providing a geometrically grounded representation for PET synthesis. The motivation is also directly supported by our results: DB-SUiT consistently outperformed deterministic surface models, alternative diffusion-bridge variants, and the volumetric baseline projected onto the cortical surface. These improvements suggest that explicitly modeling PET signals on geometrically aligned cortical surfaces better preserves cortical metabolic patterns than treating the brain purely as a volumetric image. The lobe-wise analysis further shows that the gains are spatially robust. 
The results also highlight the complementary roles of the proposed architectural and conditioning components. The SUiT backbone combines spherical convolutions for topology-preserving local feature extraction with Transformer blocks for long-range cortical interactions, which is important for modeling distributed hypometabolic patterns in dementia. The ablation study further shows that subject-level tabular information, combined additive and cross-attention conditioning, moderate surface-gradient regularization, and additional geometric surface features each contribute to improved synthesis quality.

Beyond image reconstruction, DB-SUiT-generated PET surfaces preserved diagnostically meaningful information. In the automated differential diagnosis task, synthetic PET achieved the second-best classification performance after real PET surfaces and significantly improved over MRI-based inputs. Importantly, the classifier for synthetic PET was trained on real PET surfaces, providing a stringent test of whether the generated surfaces are aligned with the real PET feature domain. 
Similar trends were observed for the CN-versus-MCI and MCI-versus-AD tasks.
These findings suggest that DB-SUiT captures disease-relevant metabolic patterns rather than only optimizing low-level reconstruction metrics.

The blinded clinical reader study provides further evidence for the clinical relevance of surface PET representations. 
The performance of the clinical reader on synthetic PET on the surface was comparable to prior results for volume-based synthetic PET \cite{li2025diffusion}. 
However, the prior volume-based method needed a clinical adaptation step to fine-tune the diffusion model on a small subset of the target dataset. 
In contrast, DB-SUiT achieved this performance by directly generalizing the ADNI-pretrained model to the unseen in-house cohort, where the ADNI dataset does not even contain FTD patients. 
We attribute this generalization performance to working on the surface data, which is more standardized and comparable across datasets than raw volume intensities. 
In addition, working on the surface reduces the dimensionality of the input and output data, yielding network architectures with fewer parameters to learn and reduced training and inference time. 



\section{Conclusion}

We introduced DB-SUiT, a novel surface-based diffusion bridge framework for MRI-to-PET translation operating directly on the cortical manifold. By introducing the conditional Spherical U-shaped vision Transformer (SUiT) with a diffusion bridge, DB-SUiT captures the complex cross-modal mapping in a geometrically grounded manner.
Across two datasets, the method achieved high-fidelity PET synthesis and preserved disease-relevant patterns, with automated diagnostic performance approaching that of real PET.
In a blinded reader study, synthetic PET yielded higher diagnostic accuracy than MRI. This result further demonstrated cross-cohort and cross-pathology generalization: the model was trained exclusively on ADNI, while the entire in-house cohort served as an external test set and included FTD cases not represented during training.
DB-SUiT may therefore provide an assistive MRI-derived estimate of cortical metabolism when PET is unavailable, while remaining complementary to clinical and cognitive information.



\section*{REFERENCES}

\bibliographystyle{IEEEtran}
\bibliography{paper}

\end{document}